\documentclass[letterpaper]{article} 
\usepackage{aaai25}  
\usepackage{times}  
\usepackage{helvet}  
\usepackage{courier}  
\usepackage[hyphens]{url}  
\usepackage{graphicx} 
\usepackage{natbib}  
\usepackage{caption} 
\usepackage{algorithm}
\usepackage{algorithmic}
\usepackage{bm}
\usepackage{amssymb}
\usepackage{amsmath}
\usepackage{booktabs}
\usepackage{graphicx}
\usepackage{subfig}
\usepackage[linesnumbered,ruled,vlined,algo2e]{algorithm2e}
\usepackage{newfloat}
\usepackage{listings}
\DeclareCaptionStyle{ruled}{labelfont=normalfont,labelsep=colon,strut=off} 
\floatstyle{ruled}
\newfloat{listing}{tb}{lst}{}
\floatname{listing}{Listing}
\title{Geometry-Aware 3D Salient Object Detection Network}
\author{
    Chen Wang\textsuperscript{\rm 1}, Liyuan Zhang\textsuperscript{\rm 1}, Le Hui\textsuperscript{\rm 1,2}\footnotemark[1], Qi Liu\textsuperscript{\rm 1}, Yuchao Dai\textsuperscript{\rm 1}\footnote{ Corresponding authors.}
}
\affiliations{
	\textsuperscript{\rm 1}Shaanxi Key Laboratory of Information Acquisition and Processing, Northwestern Polytechnical University\\
	\textsuperscript{\rm 2}PCA Lab, Key Lab of Intelligent Perception and Systems for High-Dimensional Information of Ministry of Education, Nanjing University of Science and Technology\\
	\{chenw, zhangliyuannpu\}@mail.nwpu.edu.cn,
	\{huile, liuqi, daiyuchao\}@nwpu.edu.cn\\
}

\usepackage{bibentry}

\begin{document}

\maketitle

\begin{abstract}
	Point cloud salient object detection has attracted the attention of researchers in recent years. Since existing works do not fully utilize the geometry context of 3D objects, blurry boundaries are generated when segmenting objects with complex backgrounds. In this paper, we propose a geometry-aware 3D salient object detection network that explicitly clusters points into superpoints to enhance the geometric boundaries of objects, thereby segmenting complete objects with clear boundaries. Specifically, we first propose a simple yet effective superpoint partition module to cluster points into superpoints. In order to improve the quality of superpoints, we present a point cloud class-agnostic loss to learn discriminative point features for clustering superpoints from the object. After obtaining superpoints, we then propose a geometry enhancement module that utilizes superpoint-point attention to aggregate geometric information into point features for predicting the salient map of the object with clear boundaries. Extensive experiments show that our method achieves new state-of-the-art performance on the PCSOD dataset.
\end{abstract}

%

\section{Introduction}

\label{sec:intro}
Salient object detection (SOD) focuses on segmenting the most attractive object from the surrounding background. As a pre-processing procedure, SOD has a variety of applications for many downstream tasks, such as semantic segmentation~\cite{shi2021rgb}, object detection~\cite{huang2020learning}, and visual tracking~\cite{mahadevan2012biologically}. Many efforts~\cite{fan2019shifting,liu2021learning,zhou2020gfnet} are dedicated to image-based salient object detection, and plenty of well-known works have emerged. Recently, with the rapid development of 3D sensors, such as LiDAR and Kinect camera, acquiring 3D data has become more convenient, and growing numerous point cloud based applications. However, there are few works devoted to salient object detection on 3D point clouds. Due to the irregularity and sparsity of point cloud data, it is difficult to extend the methods designed for regular 2D images to point clouds. Therefore, there are also some unresolved issues in point cloud salient object detection.

\begin{figure}[htbp]
	\centering
	\includegraphics[width=0.47\textwidth]{"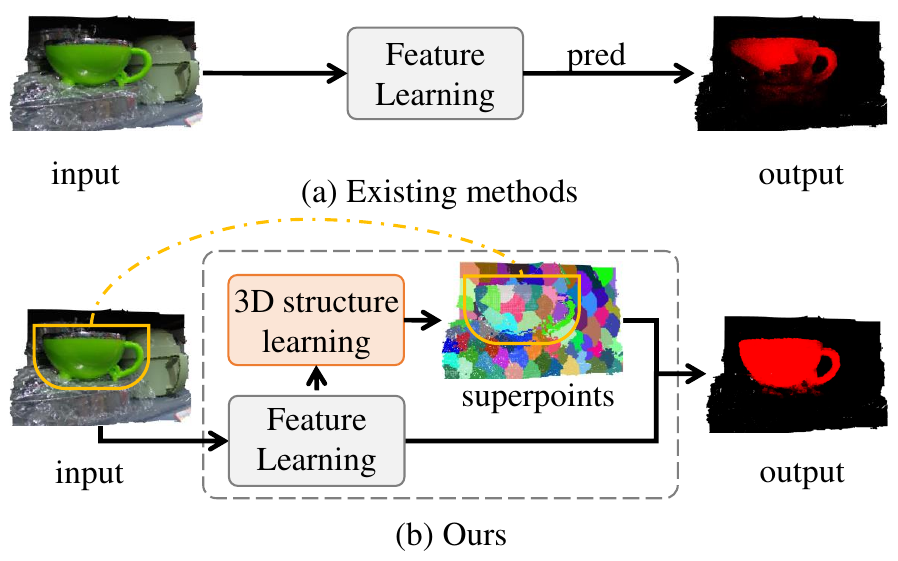"}
	\vspace{-18pt}
	
	\caption{Comparison between our method and existing methods (PointSal~\cite{fan2022salient}, EPFNet~\cite{zhang2023enhanced} and PSOD-Net~\cite{wei2024point}) in 3D salient object detection task. We explicitly utilize the structural information of the point cloud by introducing superpoint representation. It can be observed that using superpoints can effectively characterize the structure of the green cup.}
	\label{fig:mot}
	\vspace{-20pt}
\end{figure}

Many excellent works~\cite{wang2020three,liu2020dynamic,huang2020triple} have emerged salient works on images. Current state-of-the-art methods
~\cite{wang2016saliency,liu2016dhsnet,hou2017deeply,chen2018progressively,zhang2020select,tu2021multi,wang2021cgfnet} usually use multiple enhancement strategies to improve the salient object detection performance. However, accurately segmenting the object boundaries from the image still remains challenging, due to the lack of effective spatial information. Therefore, researchers considered introducing different modals for studying multi-modal salient object detection. RGB-D salient object detection~\cite{qu2017rgbd} uses depth maps to enhance the spatial information of objects for improving the accuracy of object boundaries. In addition, RGB-T salient object detection~\cite{wang2018rgb} uses RGB image combined with the thermal infrared image to locate and segment the common prominent object. Recently, researchers have begun to focus on point cloud salient object detection, $i.e.$, segmenting objects from the 3D point clouds. As a pioneer of point cloud salient object detection, Fan~\emph{et al.}~\cite{fan2022salient} proposed the first point cloud salient object detection dataset, and designed a simple multi-scale point network for 3D salient object detection task. Due to the lack of context information, it did not perform well on segmenting objects from complex backgrounds. Thus, Wei ~\emph{et al.}~\cite{wei2024point} proposed a point Transformer model to enhance contextual information of objects. Besides, in order to make full use of the complementary color information, Zhang~\emph{et al.}~\cite{zhang2023enhanced} presented an enhanced point feature network that effectively fused the RGB information with 3D point clouds. Generally, the characteristic of the human vision system is its sensitivity to the boundaries and structural information of objects. Since these 3D salient object detection methods neglect important boundary and structural information, they are unable to segment the complete boundaries of objects from complex environments (as shown in Fig.~\ref{fig:mot}(a)). Therefore, inspired by the human vision system, we consider clustering 3D point clouds into superpoints to learn the structural information of objects and fusing the superpoints with 3D points to enhance the learning of object boundaries. As depicted in Fig.~\ref{fig:mot}(b), using superpoints can effectively characterize the structural information of the object (selected by the yellow line), thereby improving segmentation performance.

In this paper, we propose a geometry-aware 3D salient object detection network, called 3DGAS, which explicitly leverages the structural information of points to enhance object boundary segmentation. The entire network consists of two parts: a superpoint partition module and a geometry enhancement module. After extracting point cloud features, we construct a simple yet efficient superpoint partition module to segment the point cloud into superpoints, a set of 3D points that share similar local geometric information. The generated superpoints are used to embed the structural information into the point features, thereby improving the accuracy of object boundary segmentation. In this procedure, the quality of superpoints determines the accuracy of object boundaries. In order to improve the quality of the superpoints, we also propose a point cloud class-agnostic loss to learn point features, which can perceive the local geometric structure of point clouds without using semantic information. After clustering superpoints, we introduce a geometry enhancement module that uses superpoint-point attention to encode structural information of the point cloud into point features, thereby strengthening the recognition ability of object boundaries.
To verify the effectiveness of the proposed method, we conduct experiments on the point cloud salient object detection dataset (PCSOD)~\cite{fan2022salient}. Extensive experiments have shown that our approach is significantly superior to other methods and has a shorter running time. We conclude the contributions as follows:

\begin{itemize}
	\item We propose a geometry-aware 3D salient object detection network, which explicitly utilizes the structural information of point clouds to enhance the segmentation of object boundaries on point clouds.
	\item We develop two simple yet efficient modules, a superpoint partition module and a geometry enhancement module, for 3D salient object detection. We also present a point cloud class-agnostic loss to learn the local geometric information of point clouds without using semantic information.
	\item Rich experiments show that the proposed method not only achieves new state-of-the-art performance on the PCSOD dataset, but also has the shortest inference time.
\end{itemize}

\section{Related Works}
\textbf{Deep learning on point clouds.} Existing deep learning based point cloud processing methods can be roughly divided into four folds: point based~\cite{lai2022stratified,jiang2020pointgroup,yan2020pointasnl,hu2020jsenet}, graph based~\cite{li2021pointvgg,ding2021graph,lei2020seggcn}, multi-view based~\cite{li2020end,xu2023multi,chen2020compositional,le2017multi} and voxelization based methods~\cite{malik2021handvoxnet++,meng2019vv,poux2019voxel}.
\begin{figure*}[t!]
	\vspace{-15pt}
	\centering
	\includegraphics[width=1.0\textwidth]{"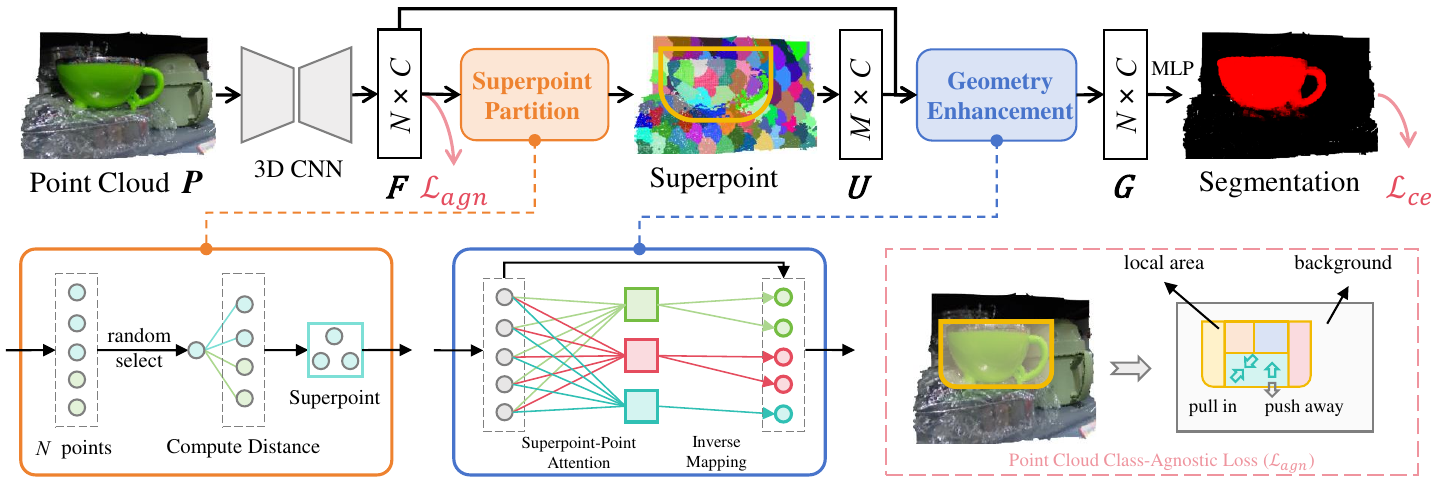"}
	\caption{The pipeline of our geometry-aware 3D salient object detection network. Given a point cloud, we first use the 3D CNN backbone to extract point features. Then, we adopt the superpoint partition module to obtain superpoints. After that, we propose the geometry enhancement module to encode structural information into point clouds. In addition, we propose a point cloud class-agnostic loss $\mathcal{L}_{agn}$ to learn discriminative point features for improving superpoint quality.}
	\label{fig:framework}
	\vspace{-10pt}
\end{figure*}
Qi~\emph{et al.}~\cite{qi2017pointnet} introduced PointNet as a pioneering method for learning features on point clouds, which directly handles point clouds with multi-layer perceptron (MLP), max-pooling, and rigid transformations to achieve extracting features of point clouds. Following PointNet, the appearance of PointNet++~\cite{qi2017pointnet++} enhanced the ability to characterize the local geometric structures of point clouds. Subsequently, in order to extends the convolution operation from 2D images to 3D point cloud, Li~\emph{et al.}~\cite{li2018pointcnn} proposed a transformation to simultaneously weight and permute the input features in PointCNN. The graph-based methods~\cite{cheng2021sspc,cheng2020cascaded,shen2018mining} regard point cloud data as a graph structure, where points represent nodes and relationships between points represent edges. These methods utilizing techniques such as graph convolutional networks to analyze and process point clouds in object recognition and segmentation tasks. For instance, DGCNN~\cite{wang2019dynamic} is a commonly used graph-based method, which dynamically aggregate local geometric feature of point clouds. The methods based voxelization~\cite{choy20194d,liu2019point} divides point cloud data into regular voxels in space. Therefore, we can use the voxelized data as input to 3D convolutional networks for further processing and analysis. These networks are specifically designed to handle three-dimensional data, allowing us to extract meaningful features and patterns from the voxelized representation. Applying 3D convolutions can capture spatial relationships and learn hierarchical representations that are useful for various tasks such as object recognition, segmentation, and reconstruction. However, the sparsity of point cloud data often results in empty voxels, leading to the wastage of computational resources. The last multi-view based method~\cite{wu2019squeezesegv2,chen2020compositional} projects point clouds into different view spaces and utilizes these views as input to accomplish analysis and processing of point clouds. Although the above methods have made significant progress in 3D classification, segmentation, reconstruction, and generation tasks, their applicability to the 3D salient object detection task is not particularly strong.

\noindent\textbf{Point cloud salient object detection.} Point cloud salient object detection refers to accurately identifying and locating salient objects from point cloud data. Similar to salient object detection in images, it often serves as a valuable preprocessing step, providing better solutions for applications such as 3D scene understanding, object recognition, and robotic navigation. However, unlike traditional image-based salient object detection, salient object detection in point clouds requires consideration of factors such as spatial distribution, density, and shape of the points. On the other hand, point cloud data consists of a large number of irregularly distributed points, posing numerous challenges for salient object detection. In recent years, with the development of deep learning techniques, significant progress has been made in salient object detection in point clouds using neural network-based methods. In point cloud SOD, Fan~\emph{et al.}~\cite{fan2022salient} proposed PointSal is the pioneering work and it take a hand-labeled dataset PCSOD for advancing this important field. In this work, PointSal is a typical encoder-decoder architecture and design two important modules, $i.e.$, point perceptron block and saliency perception block, which local salient objects though taking full advantage of multi-scale features and global semantics. However, due to PointSal capture feature all by using MLPs, the capability of learning long-range feature representations of PointSal is very limted. Subsequently, Zhang~\emph{et al.}~\cite{zhang2023enhanced} proposed an enhanced point feature network (EPFNet) for point cloud SOD, which take full advantage of color information available in point cloud for point cloud SOD. As the dominant frameworks Transform in natural language processing are applied to point clouds, Wei ~\emph{et al.}~\cite{wei2024point} later proposed PSOD-Net, a model featuring two contextual transformer modules designed to effectively capture multi-scale contextual point information.

\noindent\textbf{Superpoint representation of point cloud.} Superpoints are similar to superpixels in 2D images, which refer to a collection of points within a point cloud that exhibit certain semantic and geometric similarities. Lin~\emph{et al.}~\cite{lin2018toward} proposed a method that segments superpoints by utilizing locally crafted information to minimize an energy function. Guinard~\emph{et al.}~\cite{guinard2017weakly} utilized handcrafted local descriptors to generate geometrically simple superpoints using a greedy graph-cut algorithm~\cite{landrieu2016cut}. Landrieu~\cite{landrieu2019point} proposed employing a deep network for extracting point embeddings instead of using handcrafted features 
to segment superpoints. Hui~\emph{et al.}~\cite{hui2021superpoint} introduced an end-to-end superpoint framework, which iteratively learns the correlation mapping between individual points and superpoints for the purpose of clustering. To handle LiDAR point clouds, Hui~\emph{et al.}~\cite{hui2023efficient} propose an efficient point cloud oversegmentation network by applying clustering on the range image.

\section{Method}
\subsection{Architecture Overview}
The overall pipeline of our geometry-aware 3D salient object detection network is shown in Fig.~\ref{fig:framework}. Given a point cloud $\bm{P}\in\mathbb{R}^{N\times 6}$, where $N$ is the number of points, each point is a 6-dimensional vector that contains 3D coordinates and RGB information. We first input the point cloud into the 3D CNN backbone~\cite{yan20222dpass} to extract point cloud features $\bm{F}\in\mathbb{R}^{N\times C}$, where $C$ is the number of channels. Then, the feature $\bm{F}$ is used to calculate the point cloud class-agnostic loss for learning discriminative local geometric features without using semantic information. After that, we feed the point feature to the superpoint partition module to cluster the point cloud into $M$ superpoints. By fusing the superpoint features $\bm{U}\in\mathbb{R}^{M\times C}$ into point features $\bm{G}\in\mathbb{R}^{N\times C}$, we propose a geometry enhancement module to embed the structural information of the point cloud. Finally, we predict the mask of the object from the enhanced point features.
\subsection{Superpoint Partition}
Inspired by the human vision system, we consider introducing structural information of point clouds to improve the performance of 3D salient object detection. Specifically, we introduce the superpoint representation of the point cloud and propose a superpoint partition module to generate superpoints. The quality of superpoints determines the quality of downstream object segmentation. Therefore, before clustering points into superpoints, we propose a point cloud class-agnostic loss to help improve the quality of superpoints. Therefore, in this subsection, we only introduce the proposed simple yet effective superpoint partition approach strategy.

Similar to the concept of superpixels in 2D images, the superpoint is a set of 3D points that share similar local geometric information. To handle a large number of points, we adopt a very simple yet effective superpoint generation strategy. The core idea is to consider the similarity of point cloud features based on distance-based clustering, $i.e.$, if two points have high similarity, they belong to the same superpoint. The superpoint generation algorithm is shown in Algorithm 1. Note that in the experiment, we found that even without updating the feature of the cluster center, we can still generate good superpoints.
\begin{algorithm2e}[t]
	\caption{Superpoint Generation Algorithm}
	\label{alg:AOA}
	
	\SetKwInput{KwInput}{Input}
	\SetKwInput{KwOutput}{Output}
	
	\KwInput{Unclustered point set $U$, Queue size $K$, Distance threshold \(\gamma\)}
	\KwOutput{Superpoint sets $S$}
	
	\While{$U \neq \emptyset$}{
		Randomly select a point $i \in U$ as  cluster center\;
		Initialize queue $Q$ as empty\;
		
		\For{each point $j \in U \setminus \{i\}$}{
			Compute Euclidean distance $d_{ij}$ \;
			Enqueue $(j, d_{ij})$ into $Q$\;
		}
		Sort $Q$ by distance and keep  the $K$ nearest points\;
		
		\While{$Q$ is not empty}{
			Extract point $j$ from $Q$\;
			Compute feature distance $d(i, j)$\;
			
			\If{$d \leq \gamma$}{
				Add $j$ to the superpoint of $i$\;
				Remove $j$ from $U$\;
			}
			\Else{
				Clear queue $Q$\;
				\textbf{break}\;
			}
		}
	}
	
	\Return{$S$}\;
	
\end{algorithm2e}

\subsection{Geometry Enhancement}
After obtaining the superpoints, we use a superpoint-point attention mechanism to encode the structural information of the superpoint into the point features. 

Given the point feature $\bm{F}\in\mathbb{R}^{N\times C}$, we first obtain the superpoint feature $\bm{U}\in\mathbb{R}^{M\times C}$ by averaging the point features belonging to the same superpoint. It is worth noting that the averaging operation aggregates the local geometric information of the point cloud. Thus, the superpoint feature embeds the 3D structural information. Then, we design a superpoint-point attention mechanism to learn the correlation between the superpoints and points, which is formulated as:
\begin{equation}
	\bm{U}^{'}=\operatorname{CrossAttention}(\bm{U}, \bm{F}, \bm{F})
\end{equation}
where ``Query'' is the superpoint feature and ``Key'' and ``Value'' are point features. In this way, we fuse the fine-level point features into superpoint features. To encode superpoint features into point features, we directly inverse map superpoint features to point features based on the indexes between the points and superpoints, which is given by:
\begin{equation}
	\bm{G}=\operatorname{Inv}(\bm{U}^{'})+\bm{F}
\end{equation}
where $\operatorname{Inv}(\cdot)$ is the inverse mapping function that maps superpoint features into point features. In addition, $\bm{G}\in\mathbb{R}^{N\times C}$ is the obtained new point features. Finally, we directly use the point-level feature $\bm{G}$ to predict the category of each point, whether it belongs to the object or background.
\begin{figure*}[t!]
	\centering
	\begin{tabular}{ccccccc}
		\small View & \small GT & \small Ours & \small PSOD & \small PSal&\small PNext & \small PTrans\\

		\includegraphics[width=0.12\textwidth]{"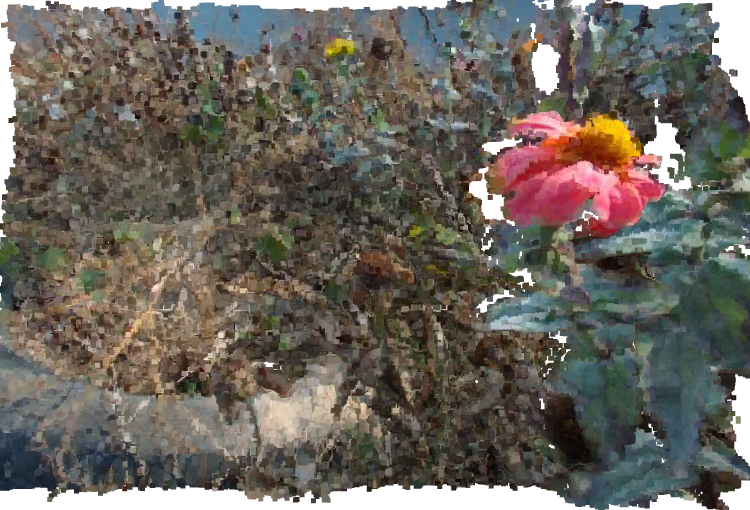"}&
		\includegraphics[width=0.12\textwidth]{"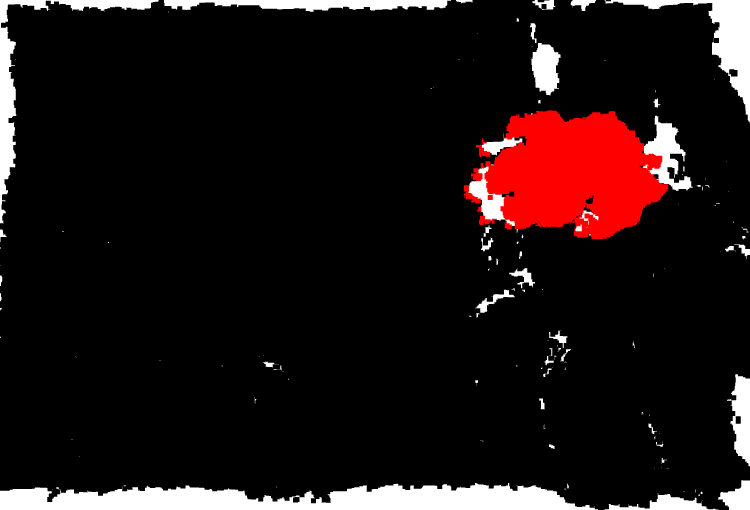"}&
		\includegraphics[width=0.12\textwidth]{"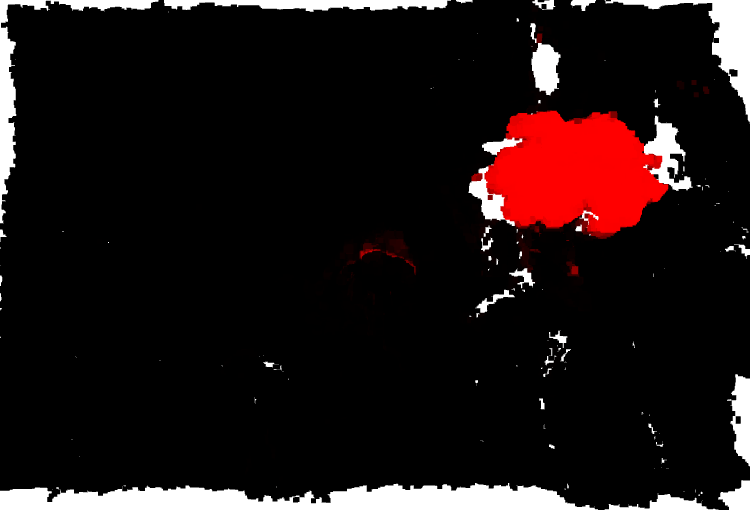"}&
		\includegraphics[width=0.12\textwidth]{"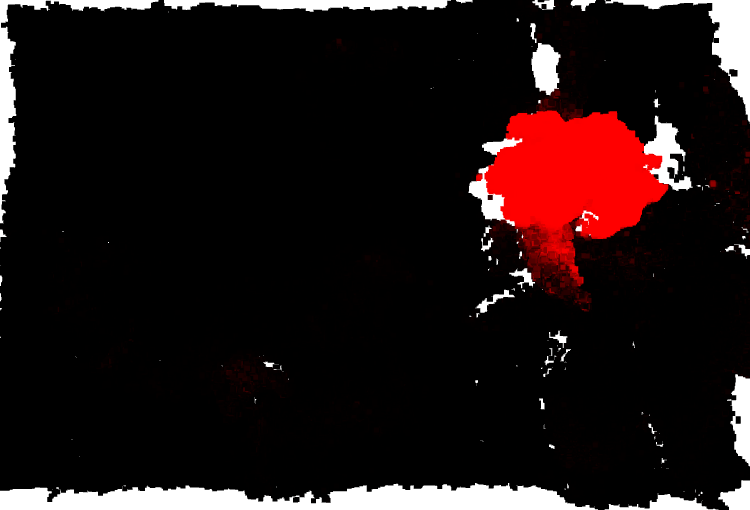"}&
		\includegraphics[width=0.12\textwidth]{"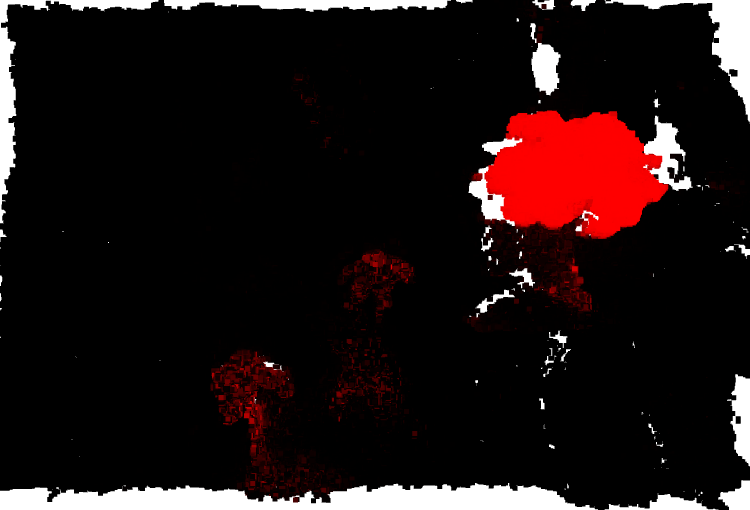"}&
		\includegraphics[width=0.12\textwidth]{"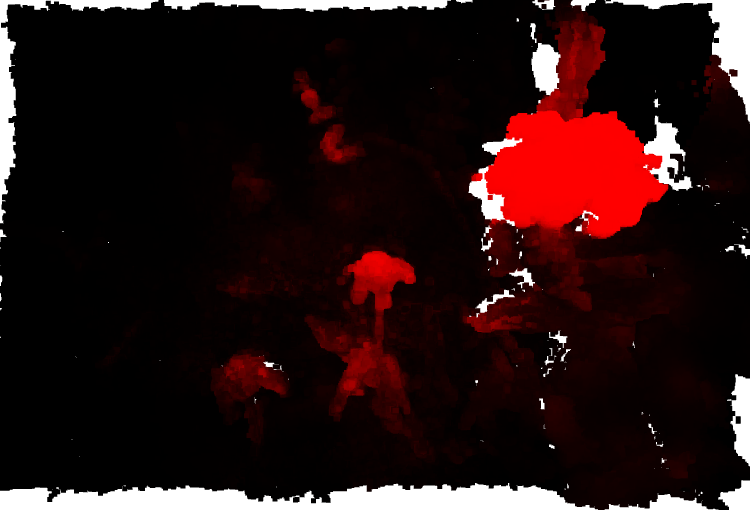"}&
		\includegraphics[width=0.12\textwidth]{"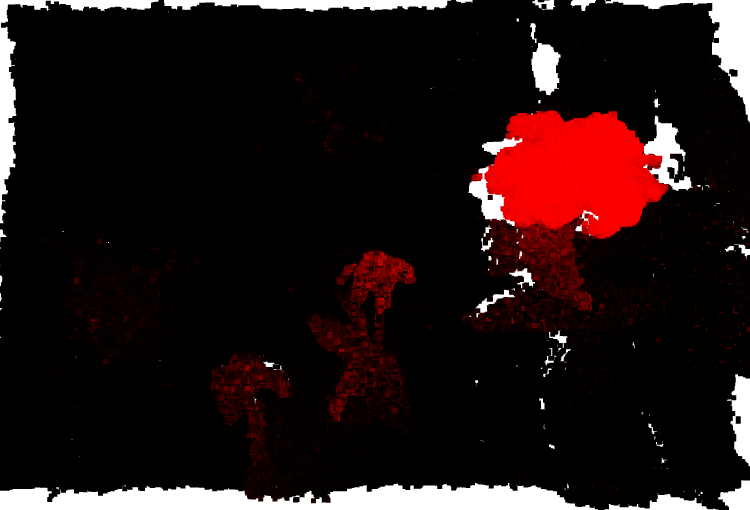"}

		\\

		\includegraphics[width=0.12\textwidth]{"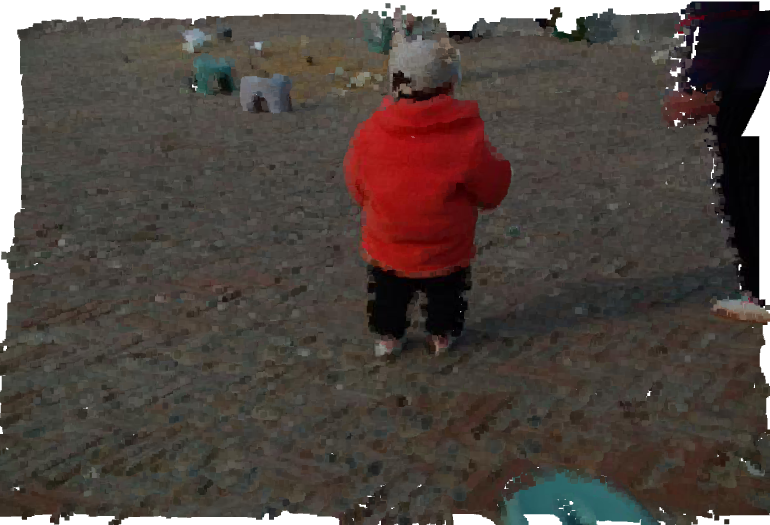"}&
		\includegraphics[width=0.12\textwidth]{"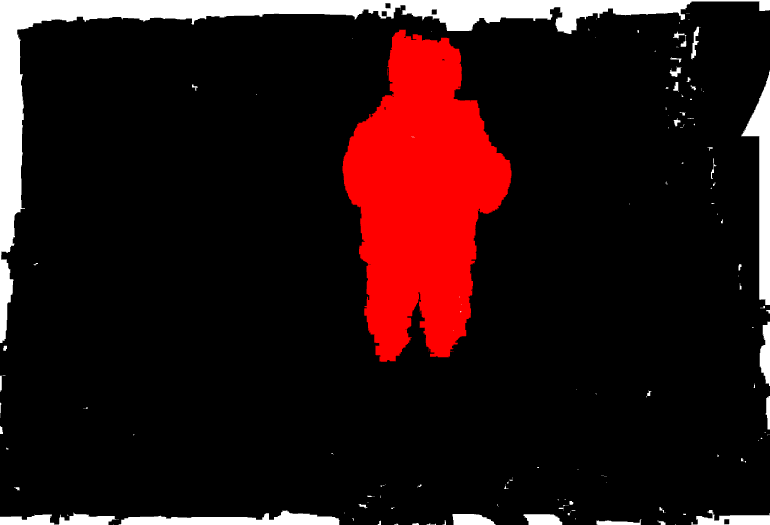"}&
		\includegraphics[width=0.12\textwidth]{"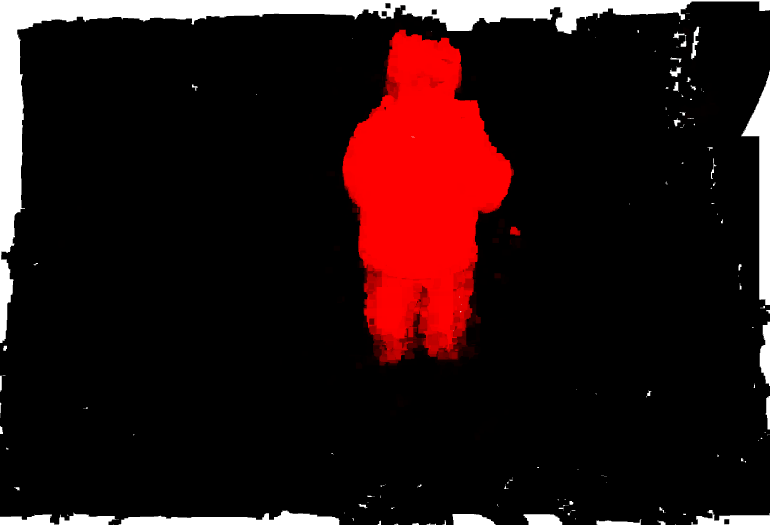"}&
		\includegraphics[width=0.12\textwidth]{"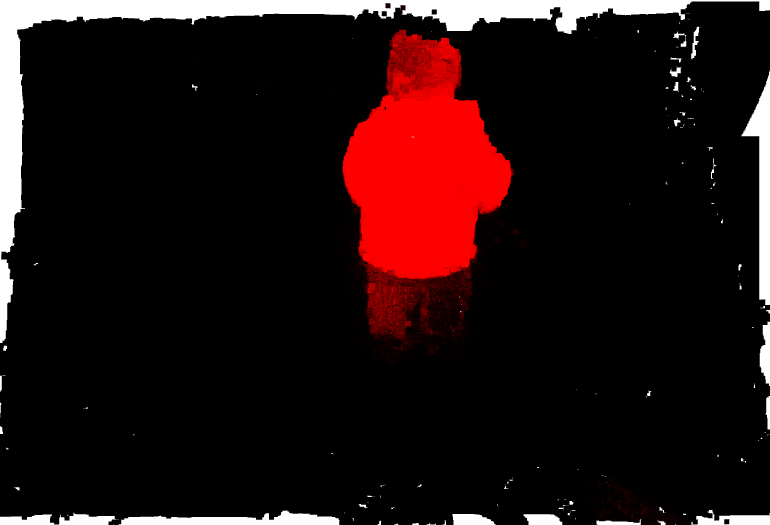"}&
		\includegraphics[width=0.12\textwidth]{"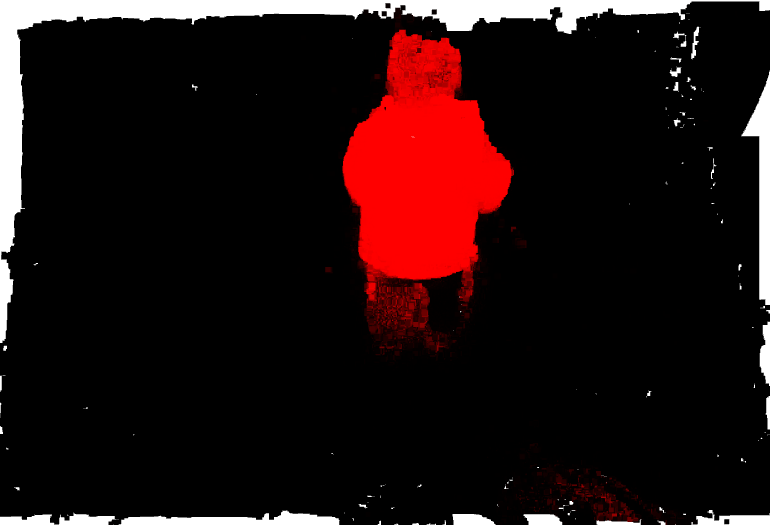"}&
		\includegraphics[width=0.12\textwidth]{"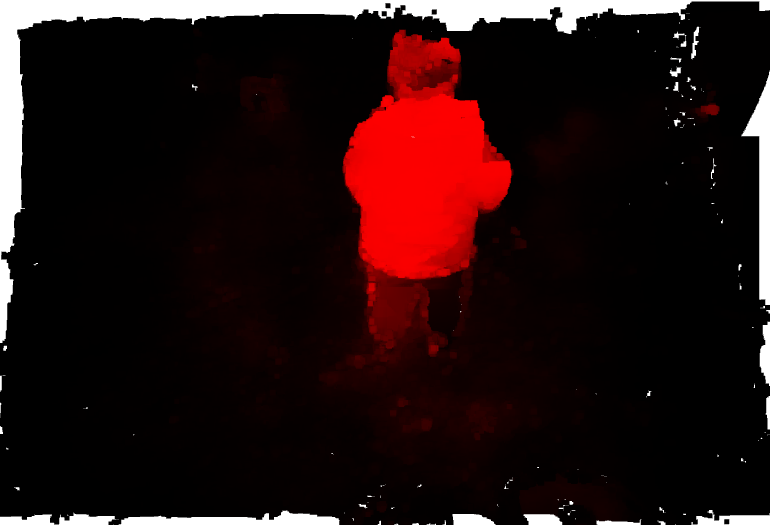"}&
		\includegraphics[width=0.12\textwidth]{"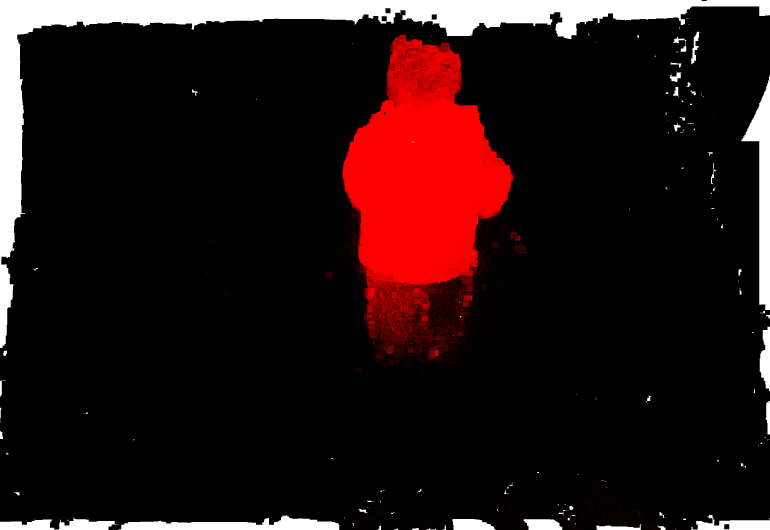"}

		\\

		\includegraphics[width=0.12\textwidth]{"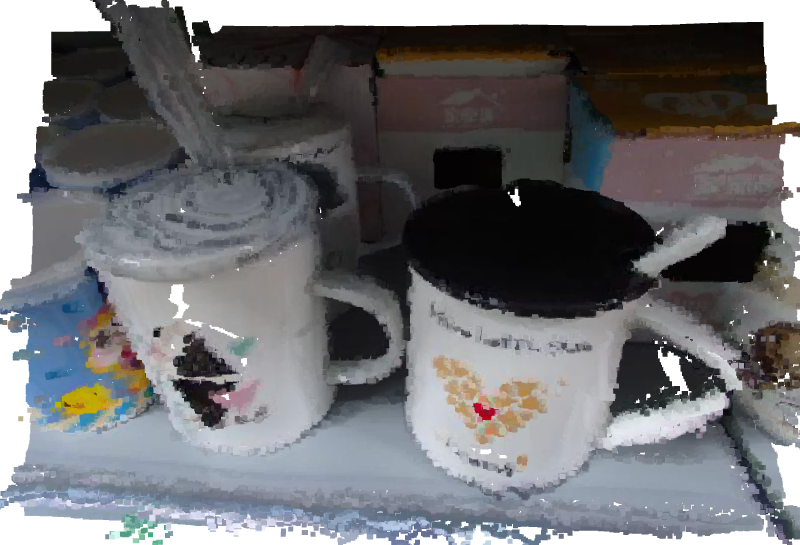"}&
		\includegraphics[width=0.12\textwidth]{"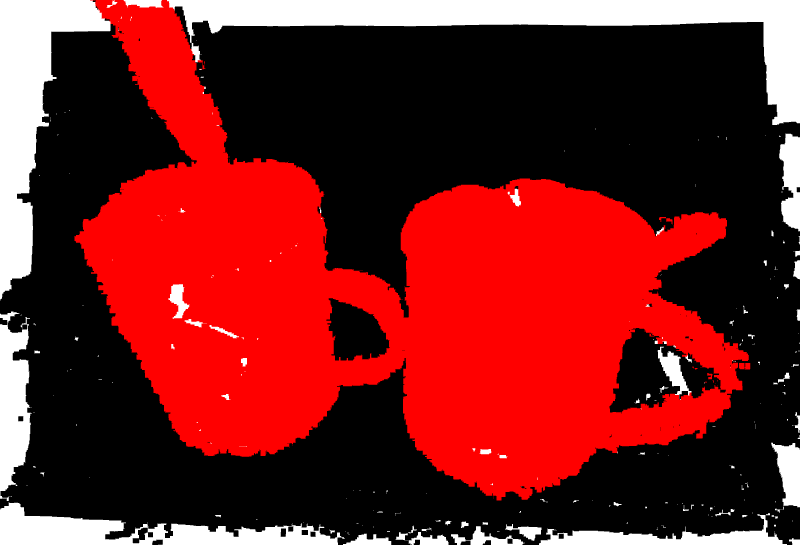"}&
		\includegraphics[width=0.12\textwidth]{"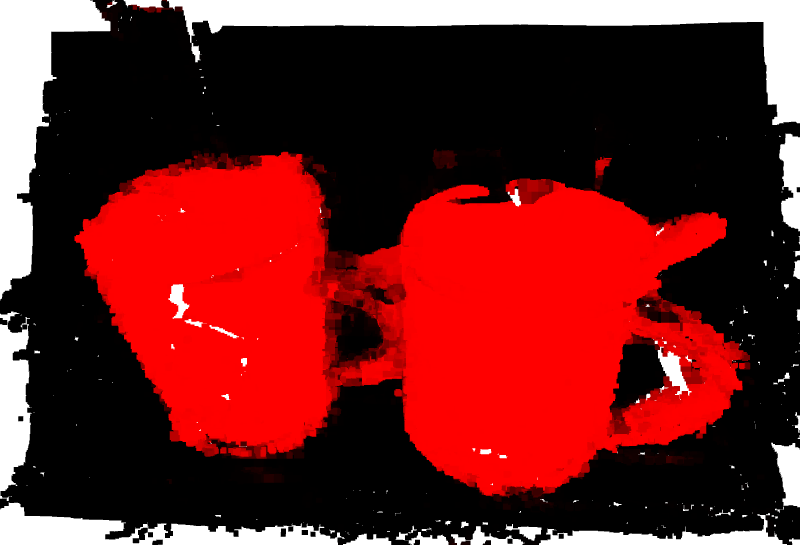"}&
		
		\includegraphics[width=0.12\textwidth]{"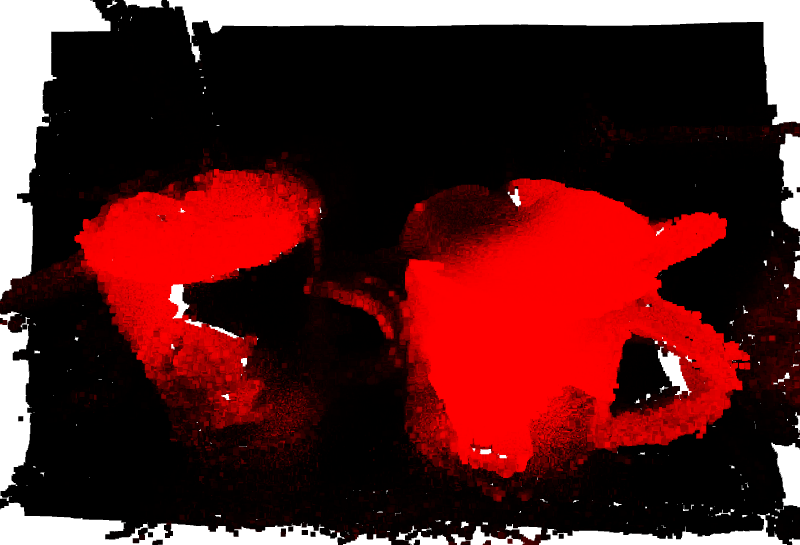"}&
		\includegraphics[width=0.12\textwidth]{"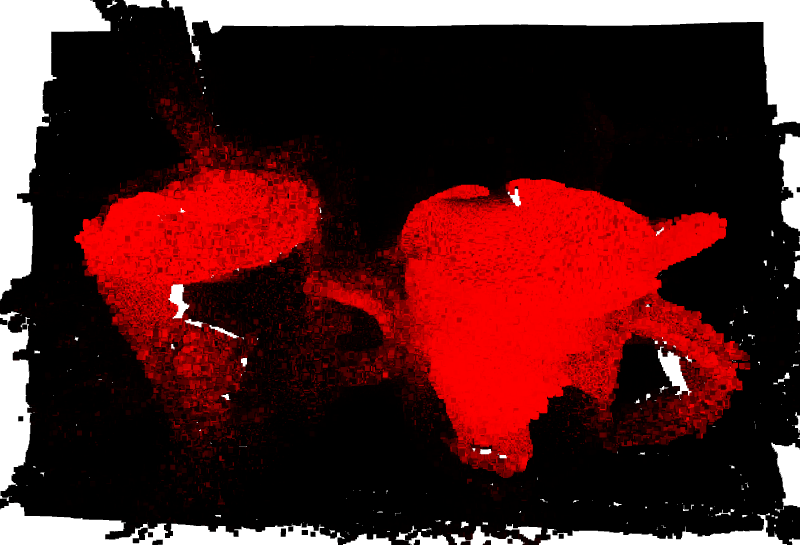"}&
		\includegraphics[width=0.12\textwidth]{"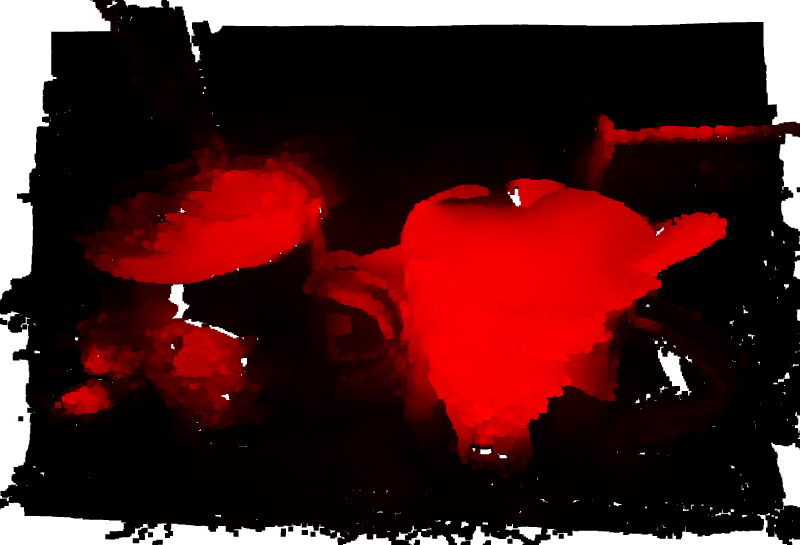"}&
		\includegraphics[width=0.12\textwidth]{"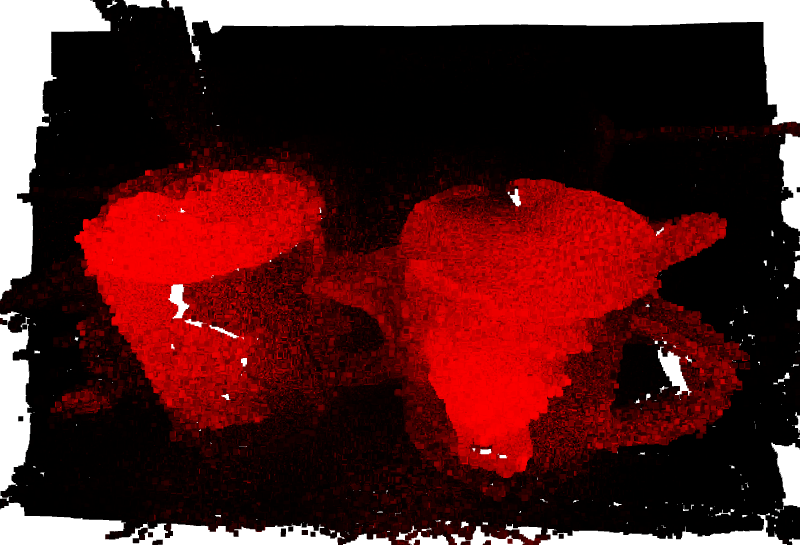"}

		\\

		\includegraphics[width=0.12\textwidth]{"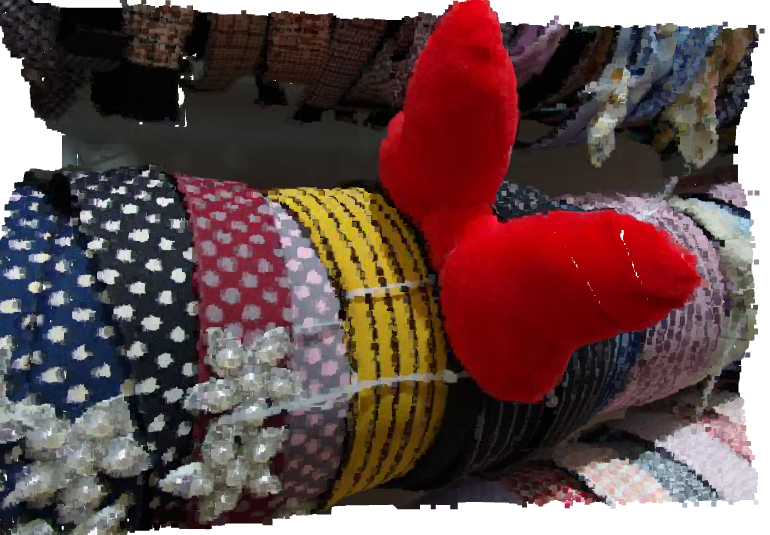"}&
		\includegraphics[width=0.12\textwidth]{"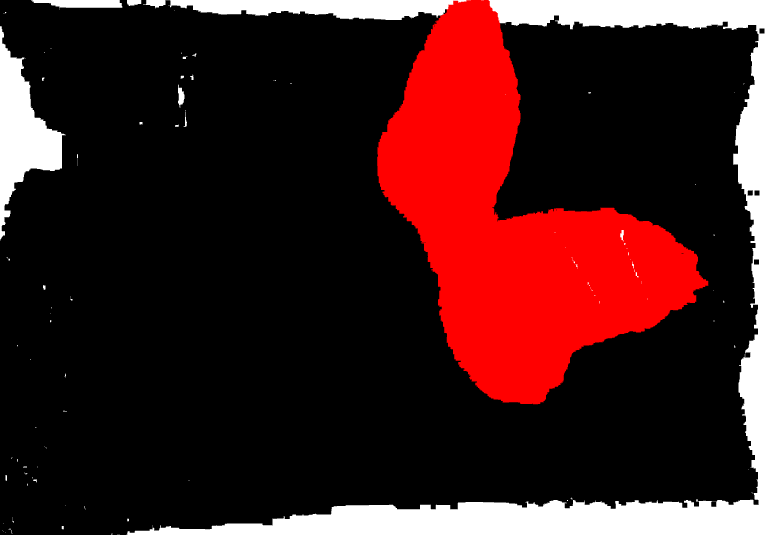"}&
		\includegraphics[width=0.12\textwidth]{"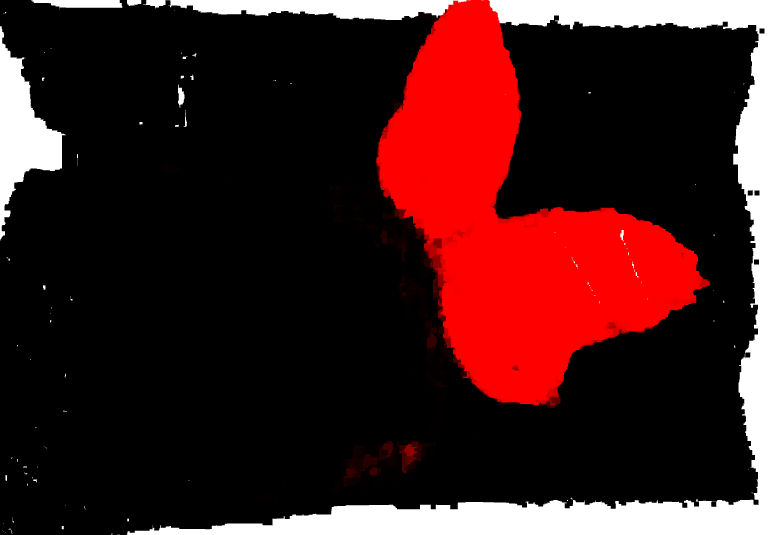"}&
		
		\includegraphics[width=0.12\textwidth]{"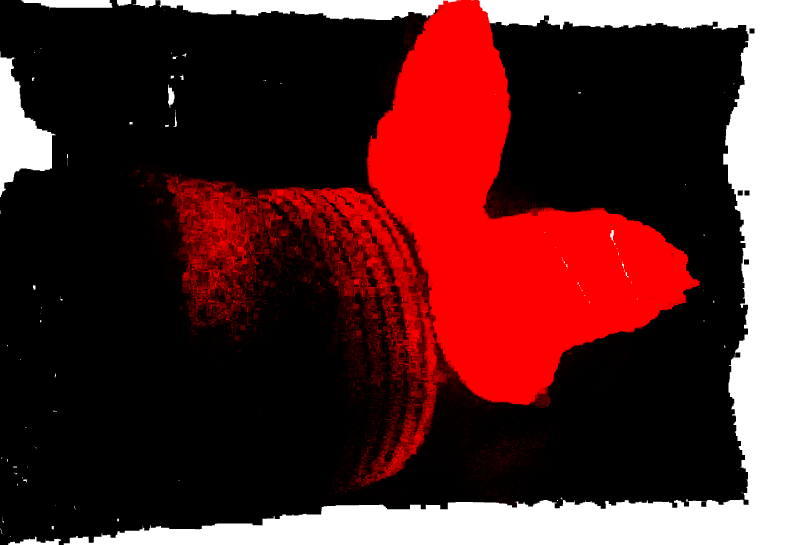"}&
		\includegraphics[width=0.12\textwidth]{"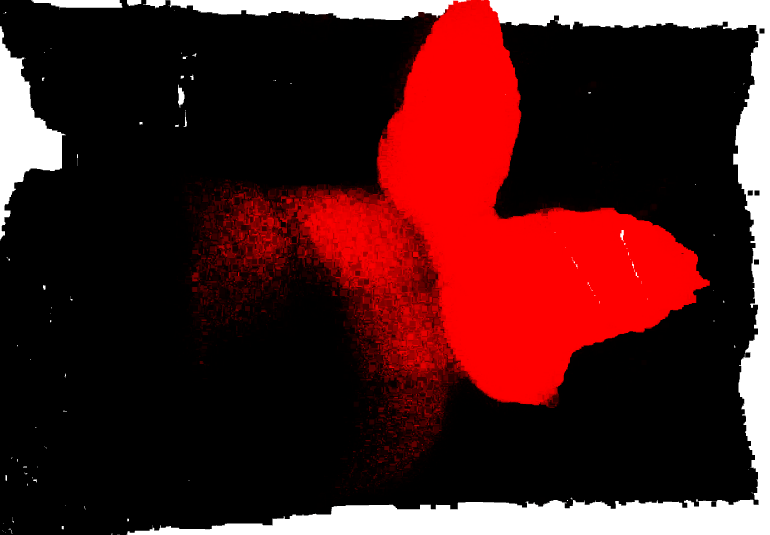"}&
		\includegraphics[width=0.12\textwidth]{"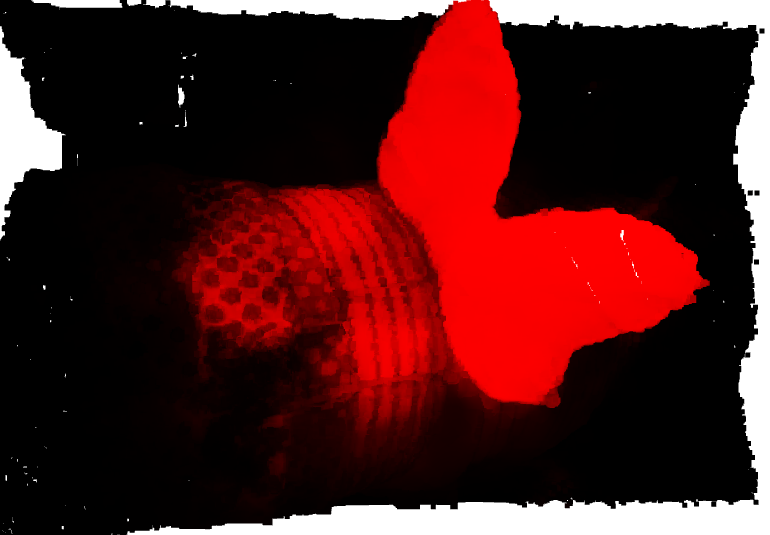"}&
		\includegraphics[width=0.12\textwidth]{"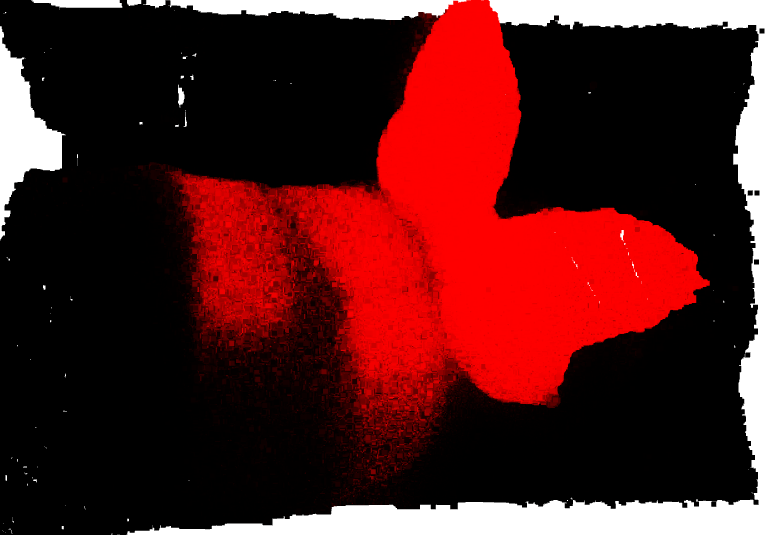"}
		
		\\

		\includegraphics[width=0.12\textwidth]{"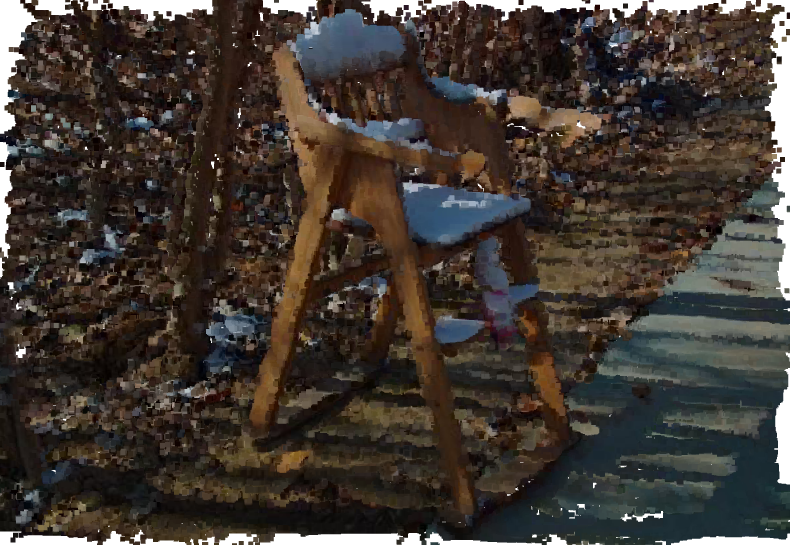"}&
		\includegraphics[width=0.12\textwidth]{"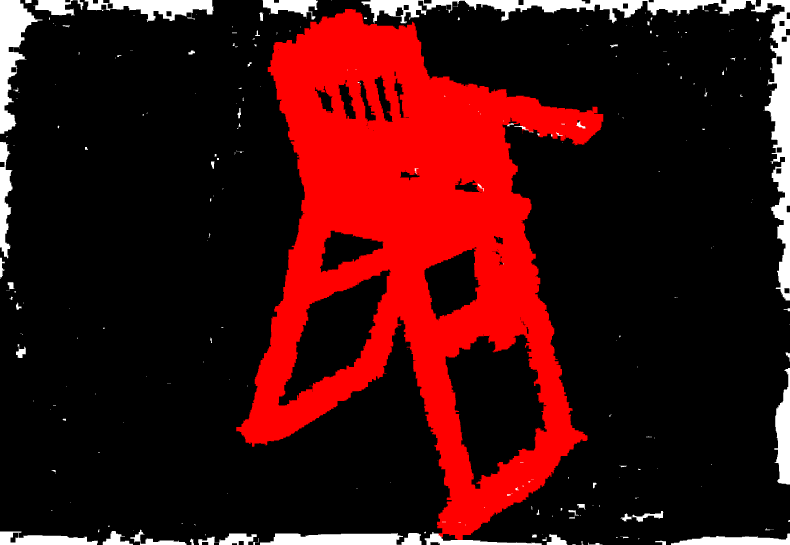"}&
		\includegraphics[width=0.12\textwidth]{"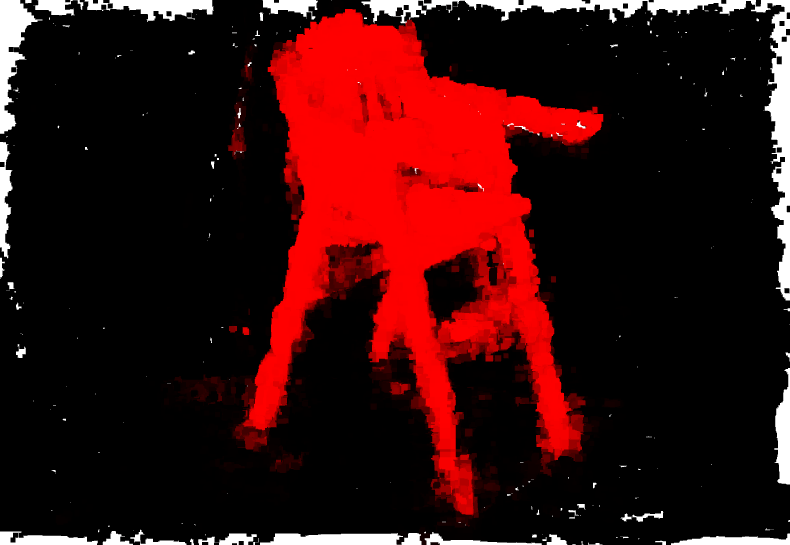"}&
		
		\includegraphics[width=0.12\textwidth]{"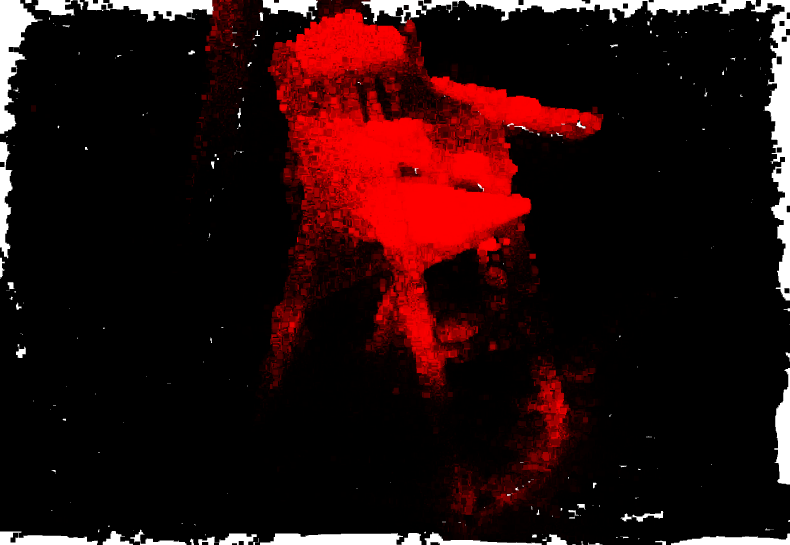"}&
		
		\includegraphics[width=0.12\textwidth]{"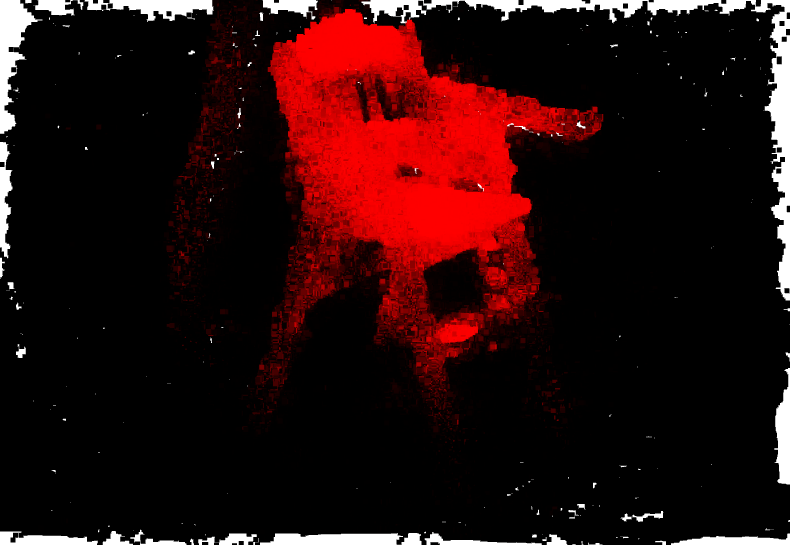"}&
		\includegraphics[width=0.12\textwidth]{"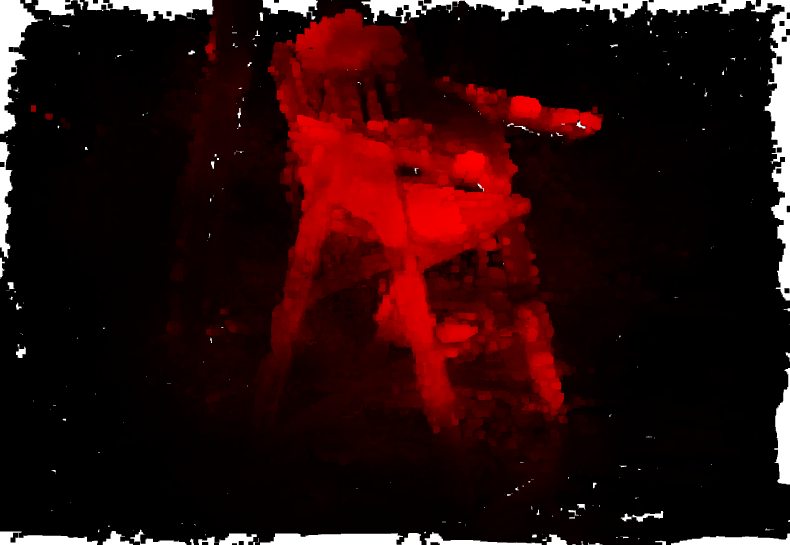"}&
		\includegraphics[width=0.12\textwidth]{"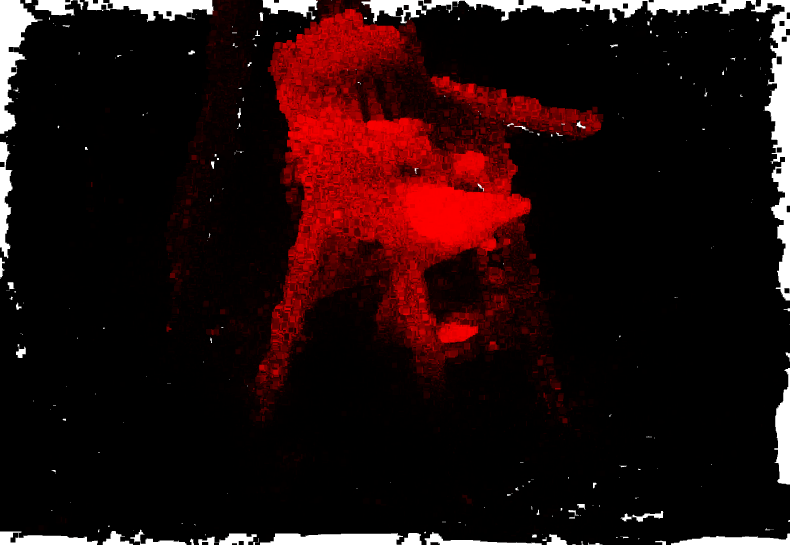"}

	\end{tabular}
	\caption{Visualization results of five methods under different views in the test set of the PCSOD dataset. Note that ``GT'',``PSOD", ``PSal'', ``PNext'', and ``PTrans'' represent the ground truth, PSOD-Net~\cite{wei2024point}, PointSal~\cite{fan2022salient}, PointNeXt~\cite{qian2022pointnext}, Point Transformer~\cite{zhao2021point}, respectively.}
	\label{fig:vis_cmp}
\end{figure*}
\subsection{Point Cloud Class-Agnostic Loss}
\label{sec:pccal}
Due to the fact that the superpoint quality is crucial for encoding geometric information into point clouds, we propose a point cloud class-agnostic loss for improving superpoint quality. In the 3D salient object detection task, each point only has a binary mask, so we cannot effectively learn semantic information from the binary mask of the point cloud. Thus, it is not possible to use semantic information to learn discriminative point features for clustering superpoints. 

Formally, the core idea of our point cloud class-agnostic loss is to consider the local area of the point cloud, rather than the overall point cloud. Compared with the whole point cloud, local areas usually have similar geometric structures. Therefore, we can utilize the similarity of local geometric structures in point clouds to constrain the similarity of local point cloud features. Specifically, given the ground truth mask of a point cloud, we first generate local areas within the object. Then, we constrain the points that are located in the same local area to have similar features to each other. Assuming that we have the $i$-th local area $\mathcal{N}_i$ within the object, we force the point $j\in\mathcal{N}_i$ close to the meaning embedding of the local area, $i.e.$, we minimizing the point feature to the meaning embedding $\bm{y}_i$ of the local area, which is given by:
\begin{equation}
	\mathcal{L}_{pull} = \frac{1}{Z}\sum\nolimits_{i=1}^{Z}\frac{1}{\left|\mathcal{N}_i\right|}\sum\nolimits_{j\in\mathcal{N}_i}\left[||\bm{f}_j-\bm{y}_i||_2-\alpha\right]_{+}^{2}
\end{equation}
where $\bm{y}_i$ is obtained by averaging the point features. $Z$ is the number of local areas. We also push the point features within the object away from the background, which is written as:
\begin{equation}
	\mathcal{L}_{push} = \frac{1}{Z}\sum\nolimits_{i=1}^{Z}\frac{1}{\left|\mathcal{N}_i\right|}\sum\nolimits_{j\in\mathcal{N}_i}\left[2\beta-||\bm{f}_j-\bm{b}||_2\right]_{+}^{2}
\end{equation}
where $\bm{b}\in\mathbb{R}^{C}$ is the background feature, which is computed by averaging the background point features. The point cloud class-agnostic loss is given by $\mathcal{L}_{agn}=\mathcal{L}_{pull}+\mathcal{L}_{push}$. Note that we empirically set $\alpha=0.01$ and $\beta=0.2$. Finally, the final loss function for training the network is written as $\mathcal{L}_{final}=\mathcal{L}_{ce}+\mathcal{L}_{agn}$, where $\mathcal{L}_{ce}$ is the cross-entropy for category prediction of each point.
\section{Experiments}
\subsection{Experimental Setup}

\hspace{1.5em}\textbf{Implementation details.} Our method is implemented with the PyTorch on an NVIDIA RTX 3090 Ti GPU. For the model, the input dims of points is 9 channels, which consist of spatial coordinates, RGB information, and normalized spatial coordinates. Since the increase in point count doesn’t significantly affect our experimental speed, we opt to utilize all 240,000 points for both the training and testing phases. For the training process, our model is trained using the Adam optimizer for a total of 300 epochs, taking approximately 13 hours. This represents a significant time saving compared to PointSal~\cite{fan2022salient} and PSOD-Net~\cite{wei2024point}, which requires training for 3000 epochs (taking roughly 3 days) and 800 epochs(taking roughly 40 hours), respectively. In our training process, in addition to using cross-entropy loss, we also introduced a point cloud class-agnostic loss to facilitate better feature learning. The initial learning rate is 1e-3 and weight decay is 1e-4. Additionally, data augmentation is applied using transformation with standard deviations of [3, 3, 3]. The parameter spatial shape is set to [150, 100, 75], indicating the shape of the spatial grid with dimensions 150, 100, and 75 along the x, y, and z axes, respectively.

\textbf{Dataset.} Following previous methods~\cite{fan2022salient,wei2024point,zhang2023enhanced}, we adopt the point cloud salient object detection dataset (PCSOD)~\cite{fan2022salient} as the benchmark to conduct experiments. PCSOD is a challenging dataset, which has 53.4\% difficult samples. It contains more than 100 daily life scenes, such as signs, daily necessities, flowers, and trees. There are a total of 2,873 scenarios in this dataset, including simple targets, multiple targets, small targets, and some targets with complex structures. According to the official partition, there are 2,000 samples for training and 872 samples for evaluation. Note that each point is described as a six-dimensional vector, which contains 3D coordinate information ($x$, $y$, $z$) and 3D color information ($r$, $g$, $b$).

\textbf{Evaluation metrics.} To assess and compare the effectiveness of various methods, we adopt the same metrics as in PointSal~\cite{fan2022salient} for performance evaluation, including mean absolute error (MAE), F-measure~\cite{margolin2014evaluate}, E-measure~\cite{fan2018enhanced}, and intersection over union (IoU). The MAE estimates the distance between the point-by-point predicted value and the corresponding true value, which is calculated as $\text{MAE} = \frac{1}{n} \sum_{i=1}^{n} |y_i-\hat{y}_i|$, where $y_i$ is the true value and $\hat{y}_i$ is the predicted value for each point. F-measure is a weighted harmonic average of precision and recall, which is formulated as $\text{F-measure}=(1-\beta^2)\cdot\frac{\text{$precision$} \cdot \text{$recall$}}{\beta^2\cdot\text{$precision$}+\text{$recall$}}$. For a fair comparison, we follow~\cite{fan2022salient} and set $\beta$=0.3 for controlling the weighting relationship between precision and recall. E-measure provides a comprehensive evaluation metric for assessing the performance of saliency segmentation models. 
It simultaneously considers both local details and global structure, thus offering a more holistic assessment of model effectiveness. 
IoU calculates the overlap ratio of predict and the truth, $i.e.$, the ratio of their intersection and union. 
For both F-measure and E-measure, we use stochastic thresholds to calculate both values to ensure that our model show excellent results at different threshold distributions.
\begin{table}[tb]

	\setlength{\tabcolsep}{4pt}  
	\centering
	\small
	\begin{tabular}{l|cccc}
		\toprule
		\multicolumn{5}{c}{Data: [xyz, rgb]} \\
		\midrule
		Methods&MAE$\downarrow$&F-measure$\uparrow$&\;F-measure$\uparrow$&IoU$\uparrow$\\
		\midrule
		
		Point Transformer & 0.075 &\;\;0.762 & 0.848 & 0.670 \\
		
		PointNeXt & 0.066 &\;\;0.779 & 0.859 & 0.680 \\
		PointSal & 0.069 &\;\;0.769 & 0.851 & 0.656 \\
		PSOD-Net & 0.058 &\;\;0.805 & 0.878 & 0.711 \\
		EPFNet & 0.047 & \;\;0.820 & 0.898 & 0.727 \\
		3DGAS (ours) & \textbf{0.042} & \textbf{\;\;0.848} & \textbf{0.912} & \textbf{0.763} \\
		\midrule
		\multicolumn{5}{c}{Data: xyz} \\
		\midrule
		
		Point Transformer$^*$ & 0.108 & 0.643 & 0.756 & 0.513 \\
		PointNeXt$^*$ & 0.092 & 0.652 & 0.752 & 0.501 \\
		PointSal$^*$ & 0.090 & 0.693 & 0.797 & 0.565 \\
		PSOD-Net$^*$ & 0.087 & 0.698 & 0.816 & 0.587 \\
		3DGAS (ours) & \textbf{0.050} & \textbf{0.814} & \textbf{0.895} & \textbf{0.720} \\
		\bottomrule
	\end{tabular}
    \caption{Comparison results of different models on the test set of the PCSOD dataset. ``[xyz,\;rgb]'' means that the color information is concatenated with the 3D coordinates. Please note that EPFNet is a multi-modal method, which uses 2D image information. ``$^*$'' means that the results are obtained by running the official codes. The best results are highlighted in \textbf{bold}.}
	\label{tab:results_pcsod}
\end{table}
\subsection{Results}
We conduct a comparative analysis of performance by comparing our method with three state-of-the-art point cloud salient object detection methods, including PointSal~\cite{fan2022salient}, EPFNet~\cite{zhang2023enhanced} and PSOD-Net~\cite{wei2024point}. We also compare with two point cloud segmentation methods, including PointNeXt~\cite{qian2022pointnext} and Point Transformer~\cite{zhao2021point}.

\begin{figure}[htbp]
	\vspace{-10pt}
	\centering
	\includegraphics[width=0.48\textwidth]{"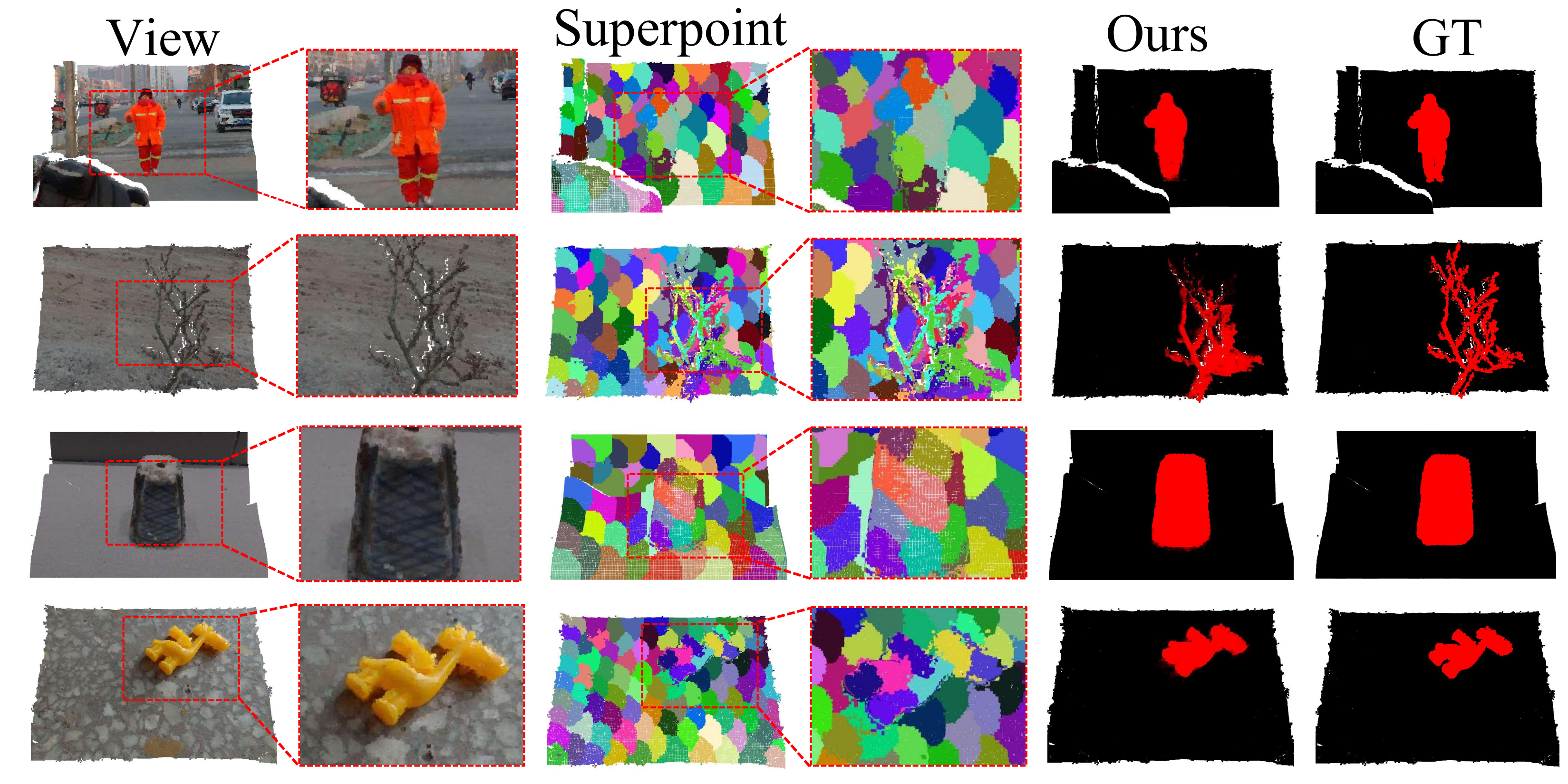"}
	\caption{Visualization results of the generated superpoints by our method in the test set of the PCSOD dataset. Please note that superpoints are randomly colored.}
	\label{fig:vis_sp}
\end{figure}
\textbf{Quantitative comparison.} We list the result of the quantitative comparison in Table~\ref{tab:results_pcsod}. Note that the PCSOD dataset contains 6-dimensional information ($xyz+rgb$). When using 3D coordinate plus color information, it can be observed that our method 3DGAS outperforms all compared methods on all four evaluation metrics. Compared with the previous state-of-the-art method EPFNet~\cite{zhang2023enhanced}, our 3DGAS surpasses it by 0.005 in MAE, 0.028 in F-measure, 0.014 in E-measure, and 0.036 in IoU. It is worth noting that except for EPFNet, all methods (Point Transformer~\cite{zhao2021point}, PointNeXt~\cite{qian2022pointnext}, PointSal~\cite{fan2022salient}, PSOD-Net~\cite{wei2024point}) concatenate 3D $xyz$ information with $rgb$ information as input. However, EPFNet is a multi-modal method that extracts color information from the point cloud as a RGB image, and uses an image network to learn features. Therefore, RGB images play an important role in the performance of EPFNet. Although our method is based on a single modal, our method can still achieve the highest performance, which further demonstrates the effectiveness of the proposed method.

To compare the learning of point cloud structures of different methods in 3D salient object detection, we conduct experiments only using the 3D coordinate information as input and discarding color information. Since the codes of  EPFNet have not been released, for a fair comparison, we conduct experiments by running the released codes of PointTransformer, PointNext, PointSal, and PSOD-Net for experiments. The quantitative results are shown in Table~\ref{tab:results_pcsod}. According to the results, our method is superior to other methods without using $rgb$ information. Even compared with the methods that uses additional color information, our method still surpasses Point Transformer, PointNext,  PointSal, and PSOD-Net. The comparison results indicate that our method can better learn the structural information of objects from the point clouds.

\begin{figure}[t]
	\centering
	\subfloat[PR curve]
	{\includegraphics[width=0.34\linewidth]{"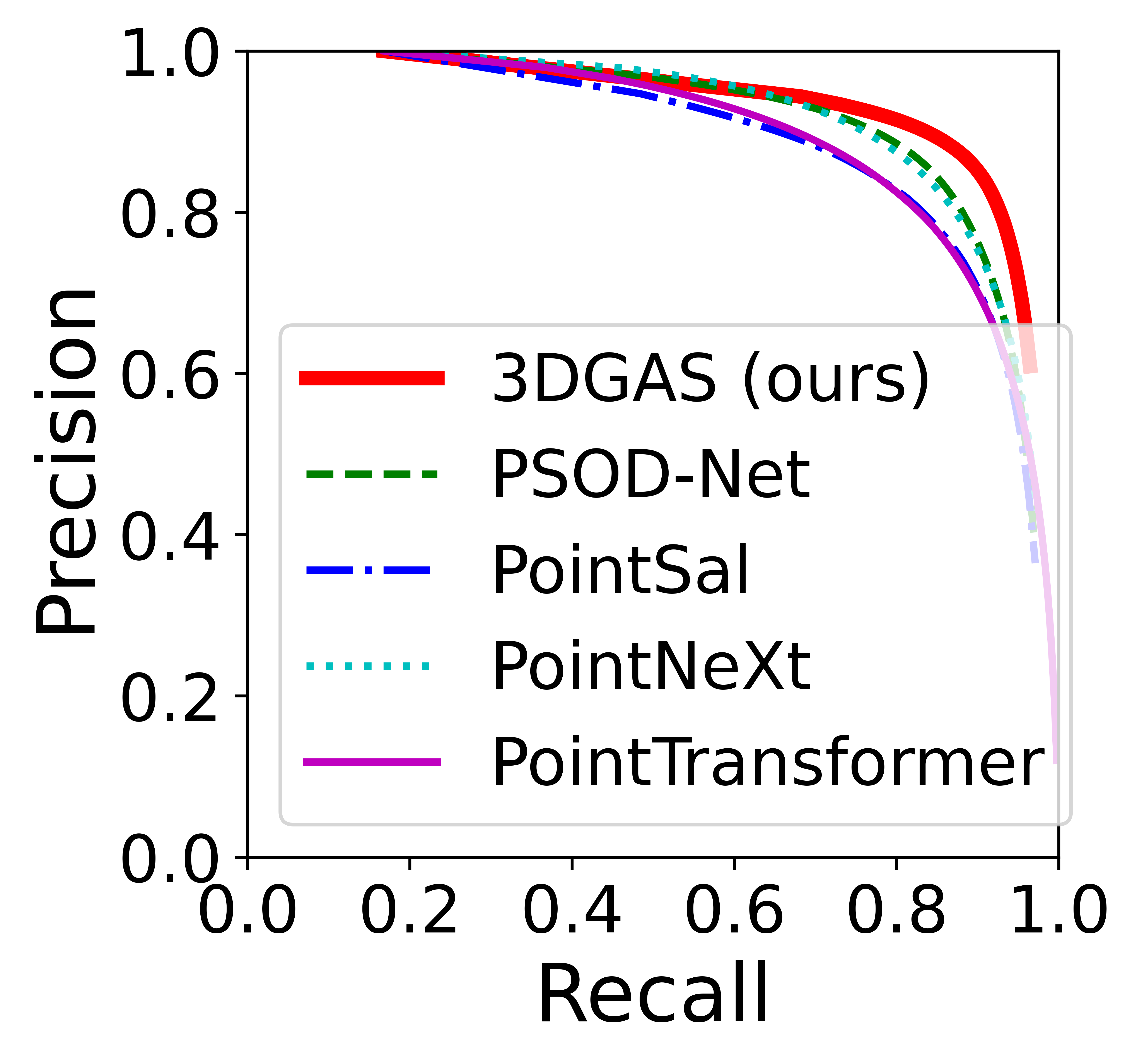"}}
	\subfloat[F-measure curve]{\includegraphics[width=.34\linewidth]{"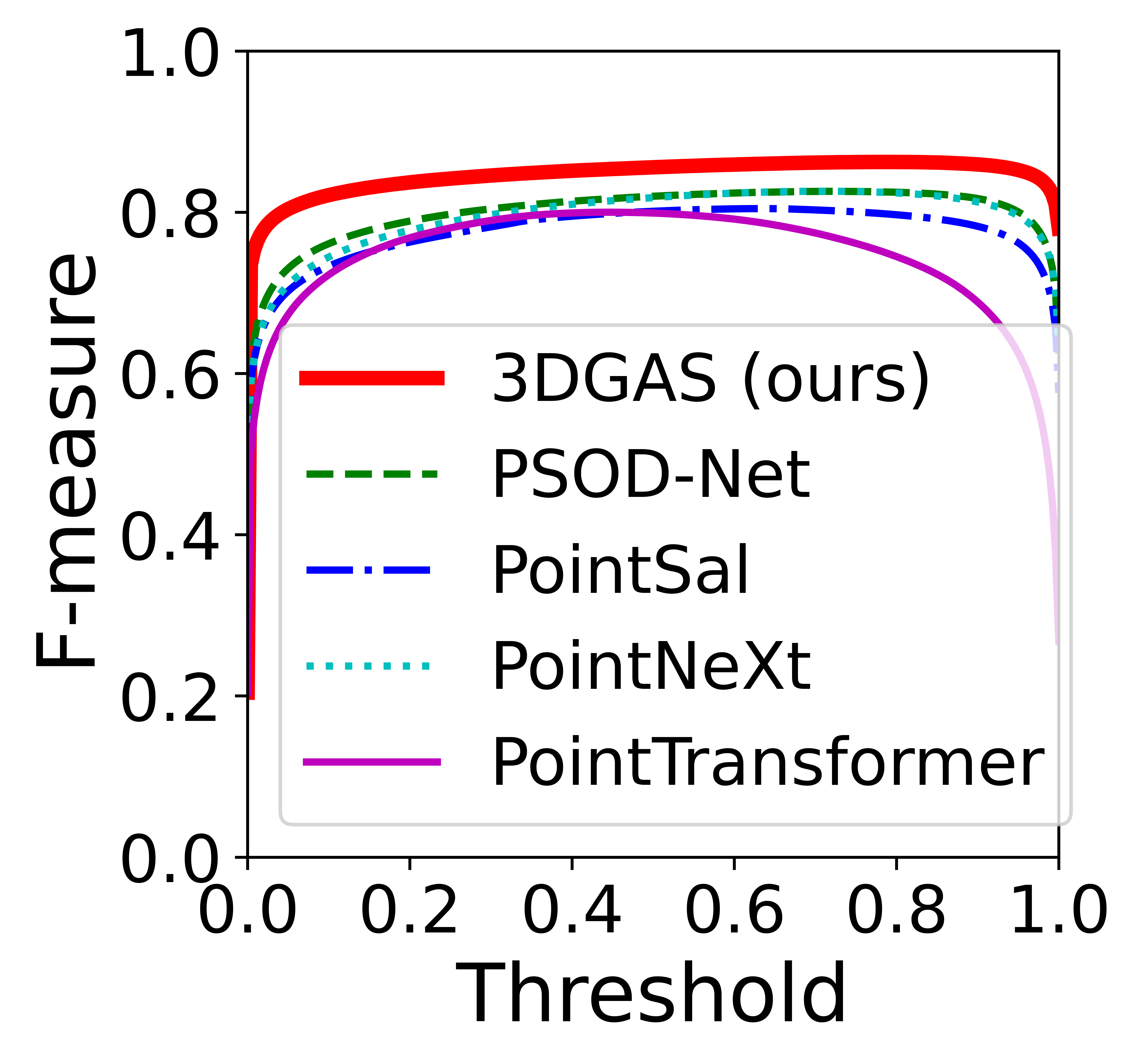"}}
	\subfloat[E-measure curve]{\includegraphics[width=.34\linewidth]{"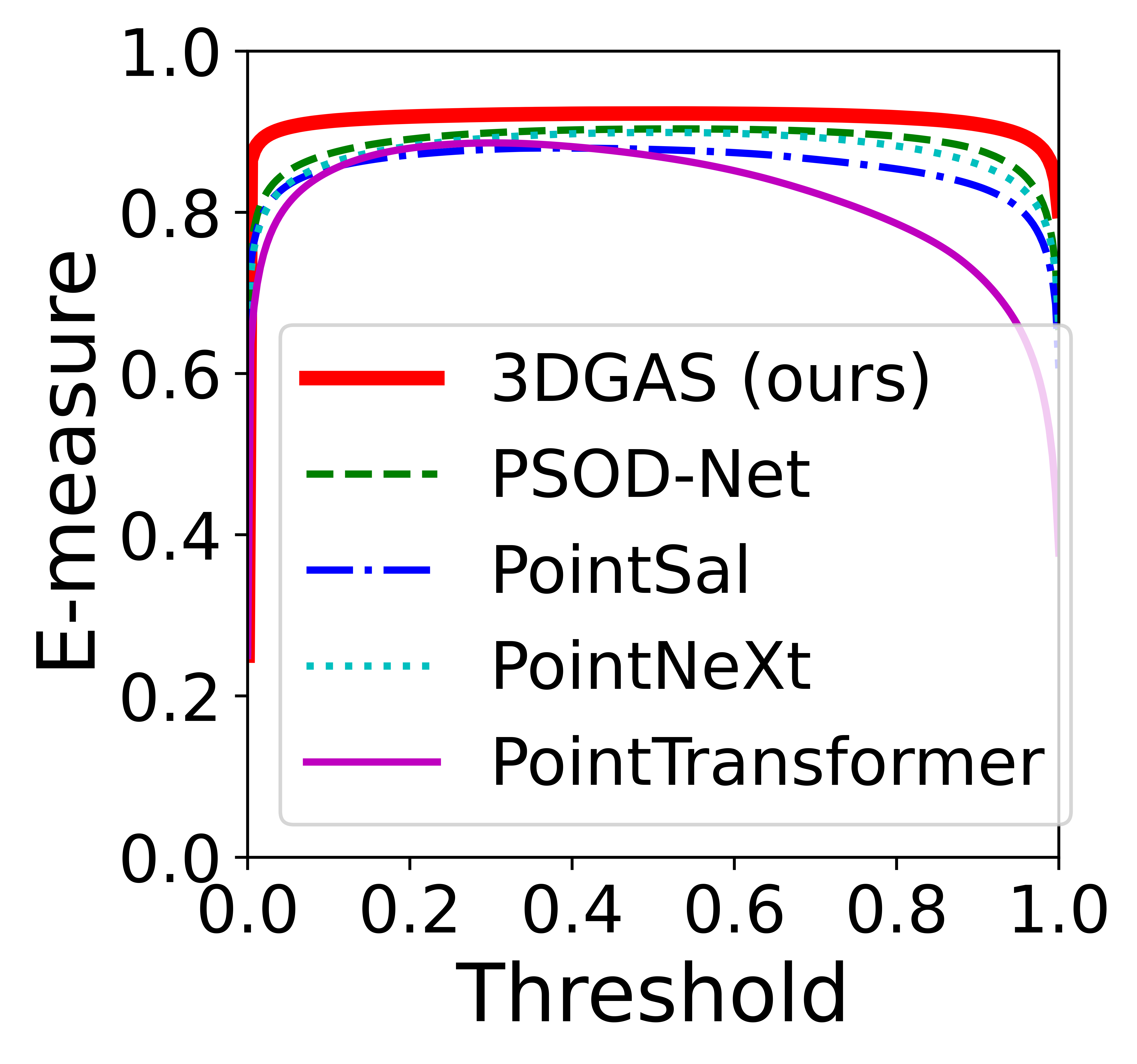"}}
	\caption{Precision-recall (PR), F-measure, and E-measure curves of different methods on the test set of the PCSOD dataset.}
	\label{fig:curves}
	\vspace{-10pt}
\end{figure}

To present the results more intuitively, in Fig.~\ref{fig:curves}, we plot the precision-recall (PR) curves, F-measure curves, and E-measure curves under different thresholds of different methods, respectively. From Fig.~\ref{fig:curves}(a), it can be found that the PR curve of our 3DGAS is higher than other methods at different thresholds, which shows that the performance of our method is better than others. According to Fig.~\ref{fig:curves}(b) and (c), it can be observed that the proposed 3DGAS consistently outperforms other methods in terms of both F-measure and E-measure at any given thresholds. Additionally, the curves of our method are smoother compared to others, indicating that our model exhibits more stable performance.

\textbf{Visualization.} To further illustrate the effectiveness of our proposed method, we present visual results of predictions on challenging scenes for different methods. As shown in Fig.~\ref{fig:vis_cmp}, the first row depicts small object detection, the second row shows the salient object detection results in simple scenes, the third row shows multi-object detection, the fourth row illustrates salient object detection in complex backgrounds, and the fifth row shows the complex structure. From the results in Fig.~\ref{fig:vis_cmp}, it can be observed that our method not only outperforms other methods in terms of localization and object segmentation integrity across various scenarios, but also exhibits favorable characteristics in edge handling. It rarely mistakes the surrounding non-target background as part of the recognition.

\begin{table}[htbp]
	
	\centering
	\setlength{\tabcolsep}{3pt} 
	\small
	\begin{tabular}{l|cccc}
		\toprule
		\ Methods & MAE$\downarrow$ & F-measure$\uparrow$ &E-measure$\uparrow$& IoU$\uparrow$ \\
		\midrule
		\ Baseline & 0.116 & 0.632 &0.768 &0.519 \\
		\ \;\;+ SP  & 0.050& 0.810& 0.881 &0.720 \\
		\ \;\;+ SP + GE & 0.047& 0.832 & 0.903 &0.738  \\
		\ \;\;+ SP + GE + CA & \textbf{0.042} &\textbf{ 0.848}& \textbf{0.912} & \textbf{0.763} \\
		\bottomrule
	\end{tabular}
    \caption{The results of ablation study on the test set of different modules. ``SP'', ``GE'', and ``CA'' indicate the superpoint partition, geometry enhancement, and point cloud class-agnostic loss, respectively.}
	\label{tab:results_ablation}
\end{table}

To reveal the contribution of the superpoints used in our method. In Fig.~\ref{fig:vis_sp}, we visualize the generated superpoints by our method in the test set of the PCSOD dataset. By comparing the visualization of view and superpoints, which are zoomed in on the second and fourth columns, it can be observed that the salient objects can be segmented into individual superpoints. For complex scenes, we can obtain good superpoints along the geometric structure of the object, such as the tree branches in the second row of Fig.~\ref{fig:vis_sp}. By introducing the structural information of point clouds through superpoints, the accuracy of boundary segmentation will be improved.

We also measure the computation costs of different methods in terms of parameters, FLOPs, and inference time. For a fair comparison, all methods are evaluated under the same experimental environment. The results are as follows: 7.1M/7.6G/0.22s (PointNeXt), 7.8M/2.8G/0.31s (PointTransformer), 4.8M/\textbf{1.4G}/1.6s (PointSal), 8.2M/4.1G/0.8s (PSOD-Net), and \textbf{0.2M}/2.6G/\textbf{0.06s} (ours). Our method achieves the lowest number of parameters due to the use of sparse convolution and lower feature channels. Although our method has higher FLOPs than PointSal, it significantly outperforms other methods in terms of performance. Additionally, by reducing the number of point clouds through voxelization, our model achieves the fastest inference time. In summary, compared to existing methods designed for 3D salient detection, our method offers faster speed and higher performance.

To evaluate the performance of point cloud salient object detection models under varying point cloud densities, we conducted a comparative experiment. Specifically, we tested PointSal, PSOD-Net, and ours on three different point cloud sizes: the original size N, half the original size N/2, and one-quarter the original size N/4. The results of MAE are present in Tabel ~\ref{tab:results_downsample}. The results indicate that our method is more robust to changes in point cloud density, making it better suited for handling point cloud data at varying resolutions in real-world applications.

\subsection{Ablation Study}
To analyze the effectiveness of each proposed module, we conduct comprehensive ablation study experiments on the PCSOD dataset.

\textbf{Effect of superpoint partition.} To analyze the effectiveness of the proposed superpoint partition, we conduct the ablation study experiment by comparing the baseline model. As shown in Table~\ref{tab:results_ablation}, when comparing the first row (baseline model) and second row, it can be observed that the proposed superpoint partition (dubbed as ``SP'') can further improve the baseline performance.
\begin{table}[t]
	
	\centering
	\setlength{\tabcolsep}{3pt} 
	\small
	\begin{tabular}{l|ccc}
		\toprule
		\ Methods & N=240,000 \;\;\; & N/2=120,000\;\;\;& N/4=60,000\;\;\; \\
		\midrule
		\ PointSal&0.069\;\;\;& 0.073\;\;\;& 0.078\;\;\;  \\
		\ PSOD-Net &0.058\;\;\;& 0.062\;\;\;& 0.086\;\;\;  \\
		
		\ 3DGAS (ours) & \textbf{0.042}\;\;\;&\textbf{0.045}\;\;\;&\textbf{0.049}\;\;\;  \\
		\bottomrule
	\end{tabular}
    \caption{MAE of different PSCOD methods under varying point cloud densities }
	\label{tab:results_downsample}
\end{table}
\textbf{Effectiveness of geometry enhancement.} Compared with previous methods, our method introduces the structural information of the point cloud through superpoints. We plus the geometry enhancement module on the learned features to conduct an ablation experiment. From Table~\ref{tab:results_ablation}, we can observe that using the geometry enhancement module (dubbed as ``GE'') can efficiently improve the performance. Due to the fact that the superpoint is a set of points that share similar local geometric information, it contributes to distinguishing the boundaries between the background and the object.

\textbf{Point cloud class-agnostic loss analysis.} In order to generate high-quality superpoints, we propose the point cloud class-agnostic loss to learn discriminative point features for clustering points. By comparing the third row and fourth row of Table~\ref{tab:results_ablation}, it can be found that using point cloud class-agnostic loss (dubbed as ``CA'') can significantly improve the performance of salient object recognition. Since it can enhance the discriminative ability of point cloud features on the local geometric structure of the salient object, the quality of generated superpoints is better.

\section{Conclusion}
In this paper, we present a geometry-aware 3D salient object detection network for point cloud salient object detection. To enhance the accuracy of object boundaries, we explicitly leverage the structural information of points by constructing superpoints. Specifically, after extracting point cloud features, we construct a simple yet efficient superpoint partition module to segment the point cloud into superpoints. The generated superpoints are used to embed the structural information into the point features, thereby improving the accuracy of object boundary segmentation.  
In order to ensure high-quality superpoints, we also propose a point cloud class-agnostic loss to learn discriminative point features for clustering points into superpoints. Extensive experiments demonstrate that the proposed method achieves new state-of-the-art performance and cost the shortest inference time.

\section*{Acknowledgments} The authors would like to thank reviewers for their detailed comments and instructive suggestions. This work was supported by the National Science Fund of China (Grant Nos. 62306238 and 62271410) and the Fundamental Research Funds for the Central Universities.

\bibliography{aaai25}

\end{document}